\documentclass{amia}
\pdfoutput=1
\usepackage{graphicx}
\usepackage{subfig}
\usepackage{xcolor}
\usepackage{amsmath, amsfonts, bm, bbm}
\usepackage{tabularx,colortbl}
\usepackage{booktabs}
\usepackage{placeins}
\usepackage[labelfont=bf]{caption}
\usepackage{hyperref}
\usepackage{natbib}
\setcitestyle{super,open={},close={}}
\begin{document}

\title{Personalized and Context-Aware Transformer Models for Predicting Post-Intervention Physiological Responses from Wearable Sensor Data}

\author{Esther Brown$^{1}$, Victoria Dean, PhD$^{2}$, Finale Doshi-Velez, PhD$^{1}$}

\institutes{
  $^1$ Harvard University, Cambridge, MA;
  $^2$ Princeton University, Princeton, NJ
}

\maketitle

\section*{Abstract}
\textit{Consumer wearables enable continuous measurement of physiological data related to stress and recovery, but turning these streams into actionable, personalized stress-management recommendations remains a challenge. In practice, users often do not know how a given intervention, defined as an activity intended to reduce stress, will affect heart rate (HR), heart rate variability (HRV), or inter-beat intervals (BBI) over the next 15 to 120 minutes. We present a framework that predicts post-intervention trajectories and the direction of change for these physiological indicators across time windows. Our methodology combines a Transformer model for multi-horizon trajectories of percent change relative to a pre-intervention baseline, direction-of-change calls (positive, negative, or neutral) at each horizon, and an empirical study using wearable sensor data overlaid with user-tagged events and interventions. This proof of concept shows that personalized post-intervention prediction is feasible. We encourage future integration into stress-management tools for personalized intervention recommendations tailored to each person's day following further validation in larger studies and, where applicable, appropriate regulatory review.}

\section*{Introduction}
Commercial wearable devices such as Garmin, Apple Watch, Fitbit, Oura, and Whoop capture dense physiological data related to stress and recovery, including heart rate (HR), heart rate variability (HRV), and beat-to-beat intervals (BBI). Yet most systems surface this data as context-free summaries (e.g., ``today's HRV'' or daily stress scores) rather than as actionable predictions that help users decide what to do next. For end users and behavioral coaches, the core questions are causal and temporal: \emph{given the context and events of my specific day, after I perform a specific intervention to alleviate stress, how will my physiology change over the next 15--120 minutes, and with what confidence?} Here, we define an intervention as a specific action or activity aimed at reducing stress.

We address this challenge by predicting post-intervention \emph{trajectories} and the \emph{direction of change} across time windows, paired with decision-aware evaluation metrics that characterize when predicted intervention effects are meaningful and reliable enough to be actionable for a user. Specifically:

\begin{enumerate}
    \item We introduce an intervention-anchored framework that uses a Transformer model to predict personalized, multi-horizon trajectories of percent change in physiological responses (BBI, HRV, and HR), expressed relative to a baseline prior to the intervention and conditioned on the user’s context.

\item We extend these personalized predictions to model the \emph{direction} (positive, negative, or neutral) of physiological change at each horizon, so the model expresses both magnitude and sign and signals when predicted effects are too small or too uncertain to be actionable for a user.

\item We apply this framework to wearable sensor data overlaid with user-tagged life events and interventions, providing a proof of concept that personalized, multi-horizon predictions of post-intervention effects on stress-related physiology are feasible. Across intervention categories and participants, we observe meaningful patterns in directional precision, physiological response profiles, and intervention heterogeneity that align with the literature on autonomic physiology.

\end{enumerate}

Our results show that future personalized stress-management tools could build on this framework to help individuals and coaches compare candidate interventions, recommending those likely to produce beneficial physiological responses for a given user in a specific context and flagging those that may work well on average but are unlikely to help that specific person.

\section*{Related Work}

\textbf{\emph{Wearable data for stress monitoring.}}
Existing work shows that signals related to autonomic state and stress,
such as heart rate (HR), heart rate variability (HRV), beat-to-beat intervals (BBI) and respiration, can differentiate rest from stress and track health status: HR and HRV data can separate resting and stress conditions~\citep{chalmers2021stress}, and resting HR data aggregate to robust clinical correlates of overall health and disease risk
~\citep{dunn2021wearable}. HRV, electrodermal activity (EDA), and respiration often outperform HR alone for stress monitoring~\citep{hickey2021smart}. Previous work also links wearable HR dynamics with stressful events~\citep{pakhomov2020using} and evaluates stress responses after mindfulness interventions~\citep{smith2020integrating,Pinge2024}. However, most of this literature focuses on detection and retrospective analysis rather than personalized, intervention-anchored predictions of physiological change that can provide actionable guidance to individuals and coaches.

\textbf{\emph{Context, self-tagging, and intervention studies.}}
Existing work has integrated user-generated context with sensor streams to improve interpretability and downstream decision-making. Some of this work prompts momentary stressor logging~\citep{neupane2024momentary} or participatory sensing with environmental context~\citep{novak2023empowering}, demonstrating value for reflection and research. Workplace just-in-time adaptive interventions (JITAIs) explore tailored intervention delivery~\citep{suh2024toward}, and randomized trials integrate wearables within structured stress-management programs~\citep{smith2020integrating}. 
However, these works primarily support retrospective analysis. In contrast, our approach \emph{predicts} post-intervention trajectories and direction of change across specific time windows, with the goal of informing personalized prospective intervention-level decisions.

\textbf{\emph{Personalization, interactive Machine Learning (ML), and human-in-the-loop.}}
Many works apply ML to personalized wearable health, spanning similarity-based prediction~\citep{comito2021diagnosis}, personality and behavior correlates~\citep{evin2022personality}, and broad surveys of personalized wearable health monitoring~\citep{olyanasab2024leveraging}. Personalization in mobile health interventions is also an active area~\citep{Brons2024,Fang2024}, while visualization systems aim to make continuous streams actionable for clinicians and patients~\citep{Sadhu2023,Arun2024}. Explainability and human-in-the-loop methods are increasingly emphasized~\citep{Abdelaal2024,Holzinger2016}, with active and interactive frameworks for wearable signals~\citep{Ashari2019} and reinforcement learning approaches informed by clinical priors~\citep{Wu2023}. 
However, few works formalize temporal, decision-aware metrics that explicitly represent and signal when predicted intervention effects are too small or too uncertain to be actionable for a user.

\textbf{\emph{Time-series forecasting, calibration, and decision-aware evaluation.}}
Prior work has shown Transformer architectures to be a powerful framework for sequence modeling and multi-horizon time-series forecasting~\citep{vaswani2017attention,lim2021temporal}, while quantile regression and related methods provide flexible ways to model predictive uncertainty~\citep{koenker1978regression}. To better reflect real-world decision-making around interventions, including both whether an intervention is likely to help and on what time scale, we focus on evaluating the predicted trajectories and their signs. In this work, we distinguish settings where the effect is positive or negative from those where it is effectively neutral, either because the predicted magnitude is near zero or because the effect is too small or too uncertain to support actionable decisions for a user. This approach is conceptually aligned with selective prediction and selective classification, where the model is allowed to abstain when confidence is low~\citep{geifman2017selective}. We tie these components together in an intervention-anchored evaluation framework and visualize the predicted effects using cluster-sorted heatmaps to reveal stable temporal sign patterns by user and intervention category.

\section*{Background}
\label{sec:background}

\textbf{\emph{Study Setting and Participants.}}
We conducted a two-phase study involving seven university students. The protocol was reviewed and approved by the Institutional Review Board at Harvard University. In Phase~1, participants wore a Garmin V\'ivosmart~5 continuously (day and night, except during charging) for two weeks. Participants had a mean age of 22.6 years (median 24, SD $\approx$ 2.9); five identified as women and two as men. In Phase~2, six participants opted into an immediate two-week extension under the same protocol. Participants recorded \emph{interventions} (e.g., meditation, walking) and routine \emph{contextual events} (e.g., classes, meetings) with start and end times using a web interface and synchronized calendar events. Across the monitoring period, participants logged 269 time-stamped tags, which we grouped into nine analysis categories, including Physical Activity (Cardio and Non-cardio), Rest \& Recovery, Food/Drink/Nutrition, Healthcare/Therapy, Socializing/Social Interaction, Spirituality/Mindful Activities, Academic \& Educational, and Other.

\textbf{\emph{Wearable Streams and User Annotations.}}
\label{sec:bg_streams}
We collected six streams from a Garmin V\'ivosmart~5—BBI, HR, steps, respiration, device measured stress, and nightly sleep score—and aligned them with user-tagged \emph{interventions} and \emph{contextual events} (names, start/end, expected effect annotations given by the user). HRV was not provided and was derived from BBI as RMSSD (rolling); see the section on ~\nameref{sec:rmssd}. HR, HRV (via RMSSD), and BBI were selected because they are widely used wearable indicators of autonomic state and stress/recovery dynamics~\citep{Shaffer2017Overview,Laborde2017HRVGuidelines,Kim2018HRVReview}. However, we do not treat these measures as definitive clinical outcomes.

\newcolumntype{Y}{>{\raggedright\arraybackslash}X}
\begin{table}[t]
\footnotesize
\setlength{\tabcolsep}{4pt}
\renewcommand{\arraystretch}{1.0}
\centering
\begin{tabularx}{\columnwidth}{@{}l c Y@{}}
\toprule
\textbf{Signal} & \textbf{Res./Unit} & \textbf{Role in stress indication} \\
\midrule
HRV (RMSSD)\footnotemark[1] & 15-min roll / 30-s step, ms &
Primary autonomic target; percent change vs.\ pre-intervention baseline; multi-horizon median and sign evaluation. \\
\cmidrule[0.2pt](lr){1-3}
Heart rate (HR) & 1-min, bpm &
Complementary arousal target; percent change vs.\ baseline; contributes to windowed sign calls and confusion matrices; interpreted with steps. \\
\cmidrule[0.2pt](lr){1-3}
Beat-to-beat interval (BBI) & per beat, ms &
Source for HRV and gap/quality checks (e.g., ectopy, gaps); not reported directly. \\
\cmidrule[0.2pt](lr){1-3}
Respiration rate & 3-min, breaths/min &
Auxiliary covariate and for stratified error analysis; not a primary target. \\
\cmidrule[0.2pt](lr){1-3}
Device stress score & 3-min, 0--100 &
Context covariate and descriptive summaries; interpreted alongside HR/HRV to avoid exercise confounds. \\
\cmidrule[0.2pt](lr){1-3}
Step count & 1-min, count &
Activity disambiguation; covariate in the forecaster; sensitivity/exclusion flags. \\
\cmidrule[0.2pt](lr){1-3}
Sleep score & nightly, 0--100 &
Prior-night context for conditioning analyses; not a target. \\
\bottomrule
\end{tabularx}
\caption{Physiological sensor variables used in this work, their sampling units, and their roles in monitoring and modeling stress and recovery after interventions.}
\label{tab:streams_roles}
\end{table}


\textbf{\emph{Intervention Effects and Time Windows.}}
\label{sec:bg_windows}
For a tagged interval with start $t_0$ and end $t_1$, we treat the intervention \emph{end} as the analysis anchor. To predict the direction of an intervention’s effect, we forecast post-intervention behavior in four windows after $t_1$: 0--15, 15--30, 30--60, and 60--120 minutes; we also report an overall aggregation of 0--120 minutes.

\subsection*{BBI and HRV (RMSSD)}
\label{sec:rmssd}
We compute heart rate variability (HRV) from the raw beat-to-beat interval (BBI) series using the root-mean-square of successive differences (RMSSD), a common choice in wearable studies.\cite{finseth2023real, jaafar2021analysis, sheridan2020heart}
Let $\Delta_i=\mathrm{BBI}_i-\mathrm{BBI}_{i-1}$ denote successive differences (ms) and $N$ the number of valid intervals in the analysis window. RMSSD is:
\[
\mathrm{RMSSD} \;=\; \sqrt{\frac{1}{N}\sum_{i=1}^{N}\Delta_i^{\,2}}.
\]
To ensure continuity, we require $N\ge 20$ intervals within a window and exclude segments containing recording gaps longer than 30 minutes (e.g., device removal or sensor dropout). RMSSD is evaluated on a rolling 15-minute window advanced every 30 seconds, then resampled to a 1-minute series by averaging the two 30-second updates per minute. This 15-minute window with 30-second stride balances short-term autonomic sensitivity with statistical stability and aligns with prior ultra-short HRV practice.\cite{Dalmeida2021HRVStress,Tervonen2021UltraShort,Velmovitsky2023AppleWatch,Liu2023DriverStress,Darwish2025ThreeStage}

\textbf{\emph{Context features.}} We compute the time-of-day and day-of-week encodings, short-horizon slopes (5- and 15-minute linear trends) for HR, RMSSD, respiration, and Garmin stress and sleep score, given by the device. These features provide context for interpreting post-intervention dynamics.

\section*{Methods}
\textbf{\emph{Methods overview.}} Our goal is to predict personalized post-intervention physiological effects across actionable horizons and to characterize when and in what direction physiology moves after an intervention. We (i) anchor analysis to the intervention end, (ii) express RMSSD, HR, and BBI as percent change from a pre-intervention baseline, and (iii) train a Transformer to predict multi-horizon trajectories and direction of effect. (iv) We then evaluate these predictions with windowed, decision-aware metrics.


\textbf{\emph{Step 1: Building Window-Anchored Examples \& Representations.}}
For each tagged interval with start $t_0$ and end $t_1$, we anchor to $t_1$ and define four post-end windows $[0,15)$, $[15,30)$, $[30,60)$, $[60,120)$ minutes, plus $[0,120)$. Signals are mapped to a 1-minute grid; HR and steps use short-gap linear interpolation, respiration and device stress use last-observation-carried-forward with a 6-minute cap (no carry across days). RMSSD is derived from BBI using a 15-minute rolling window (30-second stride; $N\!\ge\!20$ valid beats), then averaged to 1-minute resolution.

\textbf{\emph{Step 2: Learning a Multi-Horizon Forecasting Model.}}
A Transformer encoder takes as input the past $W{=}90$ minutes of multivariate context: raw streams, short-horizon slopes (5/15-minute linear trends), time-of-day/week encodings, steps/respiration/device-stress covariates, and an intervention-category embedding. Per target (RMSSD, HR, optionally BBI), quantile heads output $Q$-quantiles for each horizon; we use the median for sign evaluation. Training combines multi-horizon pinball loss, an auxiliary median MSE, an \emph{auxiliary sign loss} on the median’s sign, and a discrete-hazard head estimating return-to-baseline.

\textbf{\emph{Step 3: Train-Only Calibration \& Sign Thresholds.}}
To avoid leakage, all tuning uses train interventions only: (i) fit per-horizon isotonic calibrators on median predictions; (ii) select an onset-shift (minutes) minimizing train MAE; (iii) choose metric-specific sign thresholds $\tau_m$ that maximize train sign accuracy on \emph{non-neutral} truths after calibration. At test time we apply onset+isotonic, then call signs using $\pm\tau_m$ on predictions, while \emph{actual} neutrality uses fixed $\epsilon_m$ bands.

\textbf{\emph{Step 4: Windowed, decision-aware evaluation.}}
Within each window $[a,b]$, we evaluate the \emph{direction of change} using two related notions of accuracy. First, \emph{eligible accuracy} considers all minutes where the true sign is non-neutral (i.e., whenever the physiology clearly moved up or down). That is, minutes with $s_a(t)\neq 0$. Second, \emph{called-only accuracy} restricts the denominator to minutes where both the true sign and the model’s predicted sign are non-neutral, that is, minutes with $s_a(t)\neq 0$ \emph{and} $s_p(t)\neq 0$. We also include naive \emph{always-up} and \emph{always-down} baselines for comparison. We also use matrices (with called-only normalization) summarize TP/FP/TN/FN by metric and window to evaluate call rates.

\textbf{\emph{Step 5: Temporal Pattern Mining \& Error Forensics.}}
We summarize each intervention with a 4-window sign vector ($-1/0/+1$; NaN for missing windows) and cluster rows to produce heatmaps at (i) all-test, (ii) per-user (Figure~\ref{fig:user_heatmaps}), and (iii) per-category levels.
These visualizations surface stable temporal patterns (e.g., early HR positive signs after cardio; late RMSSD positive signs after rest) and highlight windows with calibrated abstention, signaling time ranges where the predicted intervention effects are unlikely to provide meaningful, actionable insight to a user.



\subsection*{Cohort and Data Processing}
\textbf{Participants and tagging.} As described in the ~\nameref{sec:background} section, seven university students wore a Garmin V\'ivosmart~5 continuously for two weeks; six opted for a two week extension. Participants logged \emph{interventions} and \emph{contextual events} with start/end times using a lightweight web interface, yielding 269 time-stamped tags.

\textbf{Signals and cadence.} We collected BBI (per beat), HR (1-minute), steps (1-minute), respiration (3-minute), device stress (0–100, 3-minute), and nightly sleep score (0–100). Streams were aligned to 1-minute; HR/steps used linear interpolation over short gaps; respiration/stress used capped carry-forward; sleep aligned by date.

\textbf{Leak-safe splits.} Per-user interventions were ordered by end time; the most recent block was held out as test, and earlier interventions formed train/validation. Inclusion required $\ge 24/30$ usable baseline minutes, $\ge 50\%$ valid minutes in a window, physiologic plausibility (HR 30–220\,bpm; RMSSD 1–300\,ms with clipping outside this range), and no single gap $>$ 10 minutes within a window. All normalizers, calibrators, and thresholds were fit on train only.


\subsection*{Targets and Labeling Scheme}
\label{sec:bg_targets}
We analyze post-intervention percent change for RMSSD (from BBI), HR, and BBI itself.

\paragraph{Baselines.} For metric $x_m(t)\in\{\mathrm{RMSSD},\mathrm{HR},\mathrm{BBI}\}$, the pre-intervention baseline is the 30-minute median prior to $t_0$:
\[
\mathrm{base}_m=\mathrm{median}\!\big(x_m[t_0-30,\,t_0)\big).
\]

\paragraph{Percent change and signs.} For $t\ge t_1$,
\[
\Delta\%_m(t)=100\cdot\frac{x_m(t)-\mathrm{base}_m}{\max\big(|\mathrm{base}_m|,10^{-6}\big)}.
\]
Minute-level ground-truth signs use metric-specific neutrality bands:
\[
s_a(t)=
\begin{cases}
+1 & \Delta\%_m(t)\ge \epsilon_m,\\
-1 & \Delta\%_m(t)\le -\epsilon_m,\\
0  & \text{otherwise,}
\end{cases}
\quad
\epsilon_{\mathrm{RMSSD}}{=}2.5,\;\epsilon_{\mathrm{HR}}{=}1.0,\;\epsilon_{\mathrm{BBI}}{=}1.0 \;\text{(pp).}
\]
These labels support direction-of-change evaluation within each post-intervention window.


\section*{Results}

\begin{figure}[!htb]\centering
\subfloat[\textbf{Called-only windowed sign accuracy (User G).}
Panels/columns are ordered \textbf{left}\,$\to$\,RMSSD, \textbf{middle}\,$\to$\,BBI, \textbf{right}\,$\to$\,Daily Heart Rate.]{
  \includegraphics[width=.85\linewidth]{img/prediction/user_g_test_interventions.pdf}
  \label{fig:user_g_test_interventions}
}\\[-2pt]
\subfloat[\textbf{Confusion matrices by metric and window (User G; Percentage of called points).}
Panels/columns use the same left→middle→right ordering: RMSSD, BBI, Daily Heart Rate.]{
  \includegraphics[width=.96\linewidth]{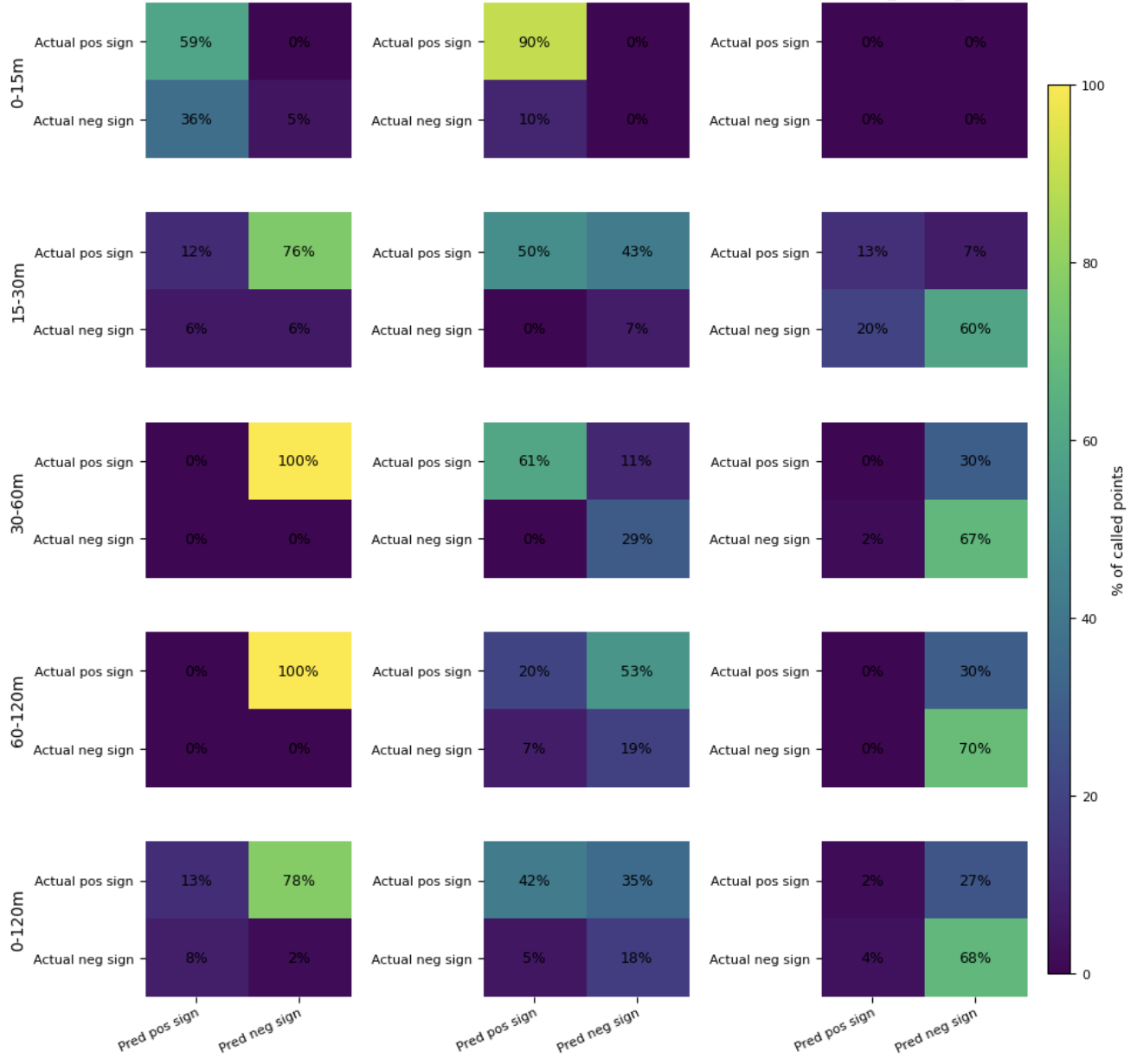}
  \label{fig:confusion_matrix_g}
}

\caption{\textbf{User G — called-only accuracy and error composition (same panel ordering across plots).}
\textbf{(a)} Bars report the percent of correct \emph{direction-of-change} calls on \emph{called minutes} within post-end windows (0–15, 15–30, 30–60, 60–120\,min); the column to the right of the red dashed divider summarizes 0–120\,min overall. 
\textbf{(b)} Confusion matrices decompose those calls by window; rows are actual (positive/negative), columns are predicted (positive/negative), and cell values are the percentage of \emph{called} points. For each metric and window, the diagonal sum matches the corresponding bar height in (a) up to rounding. 
\textbf{Takeaway:} BBI shows the strongest and most consistent called-only accuracy across windows; HR is moderately accurate (predominantly negative signs), while RMSSD carries useful signal mainly in the first 15 minutes.}
\label{fig:userG_called}
\end{figure}

\begin{figure}[!htb]
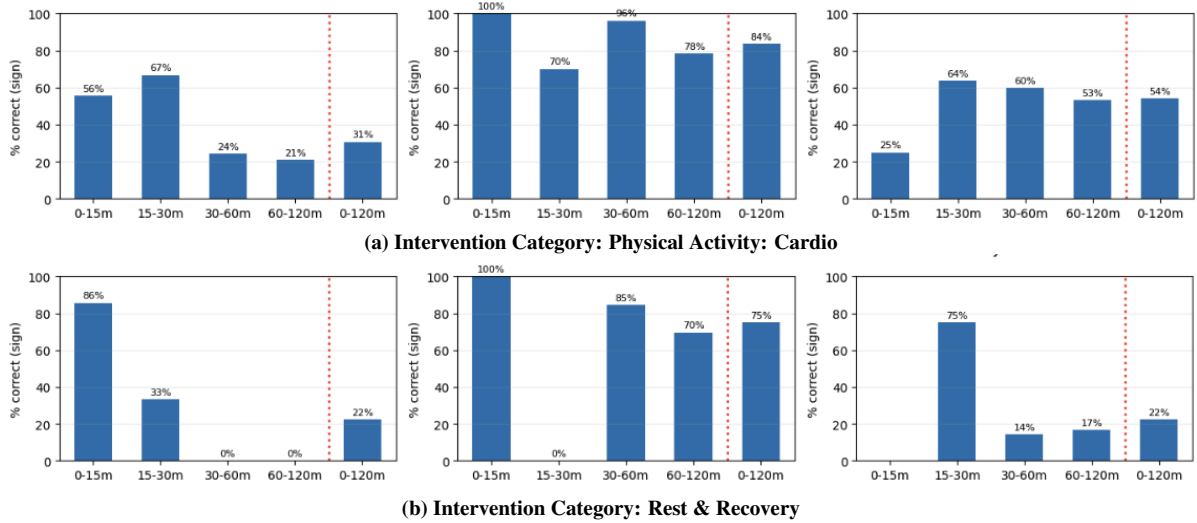
\centering
\subfloat[\textbf{Intervention Category: Physical Activity: Cardio}]{
  \includegraphics[width=.96\linewidth]{img/prediction/physical_activity_test.pdf}
  \label{fig:category_cardio_called}
}\\[-2pt]
\subfloat[\textbf{Intervention Category: Rest \& Recovery}]{
  \includegraphics[width=.96\linewidth]{img/prediction/rest_recovery_test.pdf}
  \label{fig:category_rest_called}
}
\caption{\textbf{Category-level windowed sign accuracy (non-neutral points).}
Bars show the percent of correct \emph{direction-of-change} calls within post-end windows
(0–15, 15–30, 30–60, 60–120\,min); the bar to the right of the red dashed divider summarizes
0–120\,min overall. Panels are ordered left$\to$RMSSD, middle$\to$BBI, right$\to$Daily Heart Rate.
\textbf{Takeaways:}  
(a) Cardio: BBI is consistently strong across windows (100\%, 70\%, 96\%, 78\%; overall \textbf{84\%}); HR is moderate beyond the first 15 minutes (64\%, 60\%, 53\%; overall \textbf{54\%}); RMSSD is mainly early (56\%, 67\%) and weak thereafter (24\%, 21\%; overall \textbf{31\%}). 
(b) Rest \& Recovery: RMSSD is strong early (86\% at 0–15\,min) then fades; BBI is very strong at 0–15\,min (100\%) and again at 30–60\,min (85\%), yielding \textbf{75\%} overall; HR shows a brief 15–30\,min effect (75\%) and is otherwise weak.}
\label{fig:category_cardio_rest}
\end{figure}

\textbf{\emph{Short-horizon direction is predictable, and different physiological features provide different levels of signal.}} 
Across participants and intervention categories, beat-to-beat intervals (BBI) provide the most consistent and stable windowed direction accuracy within the first 60 minutes after an intervention (Figures~\ref{fig:userG_called},~\ref{fig:category_cardio_called},~\ref{fig:category_rest_called}). This pattern aligns with the literature on autonomic physiology: BBI reflects the instantaneous timing between heartbeats and responds rapidly to both sympathetic and parasympathetic shifts~\citep{shaferHRVBioReview2017,kimHRVReview2018}. Because these beat-to-beat changes can occur on a sub-second scale, BBI captures short-horizon autonomic adjustments more reliably than aggregated measures such as daily heart rate~\citep{Shaffer2017Overview,laborde2017hrv,stanleyHRvsHRV2013,buchheit2014sensitivity}. In the context of our motivating question—\emph{given the context and events of my specific day, after I perform a specific intervention, how will my physiology change over the next 15–120 minutes?}—we observe that BBI offers the strongest signal for making personalized, intervention-level predictions of direction of change over short horizons.

RMSSD, in contrast, primarily reflects short-term parasympathetic (vagal) activity—the branch of the autonomic nervous system responsible for slowing the heart and supporting recovery. Its strongest signal appears immediately after an intervention, typically within the first five to fifteen minutes, after which the metric often plateaus or becomes noisy~\citep{labordeHRVGuidelines2017,jarvela_earlywindow2021}. Consistent with this physiology, RMSSD predictions in our framework show meaningful accuracy only in the earliest post-end window, with diminished signal thereafter (Figures~\ref{fig:userG_called},~\ref{fig:category_cardio_called},~\ref{fig:category_rest_called}). This suggests that RMSSD is particularly useful for answering how physiology changes for a given user in the first 15 minutes after an intervention, but less reliable for longer-term forecasts.

We also observe that daily heart rate (HR) exhibits moderate accuracy and often trends negative following activity-related interventions (Figure~\ref{fig:userG_called}). This aligns with the known slower kinetics of HR recovery and its sensitivity to non-autonomic influences such as movement, metabolic load, and thermoregulation~\citep{bruceHRRecovery1996,stanleyHRvsHRV2013}. As a result, HR direction can be predictable over short windows but is less granular than BBI and less temporally focal than RMSSD when predicting the effects of interventions. For our central question, HR is better suited to supporting broader, per-user guidance (e.g., overall downward trends after activity) and is not the primary signal for fine-grained personalized predictions about the effects of an intervention within the context of a given day.

\textbf{\emph{Directional accuracy is stable where the signal is strong and confusion patterns are physiologically consistent.}}
Windowed confusion matrices (Fig.~\ref{fig:confusion_matrix_g}) show that the composition of direction-of-effect calls closely mirrors the physiological prediction patterns observed across features. For BBI, early windows contain substantial proportions of both true-positive and true-negative calls, reflecting strong short-horizon autonomic responsiveness and rapid sensitivity to sympathetic and parasympathetic shifts~\citep{shaferHRVBioReview2017,kimHRVReview2018}. In contrast, RMSSD’s confusion matrices in later windows show high true-negative rates and very low call rates, consistent with RMSSD’s brief early parasympathetic (vagal) sensitivity and its well-documented decline beyond the first 15 minutes after an intervention~\citep{labordeHRVGuidelines2017,jarvela_earlywindow2021}. HR exhibits mostly true-negative calls with sparse positive predictions, aligning with its slower recovery kinetics and its dependence on non-autonomic factors such as movement and thermoregulation~\citep{bruceHRRecovery1996,stanleyHRvsHRV2013}. Taken together, these matrices demonstrate that the model makes confident direction predictions when the underlying physiological signal is strong and becomes conservative (often defaulting to no or negative calls) when the signal is weak.

\begin{figure}[ht]
\includegraphics[width=.46\linewidth]{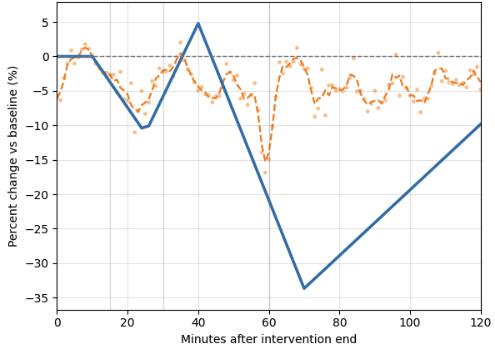}\hfill
\includegraphics[width=.46\linewidth]{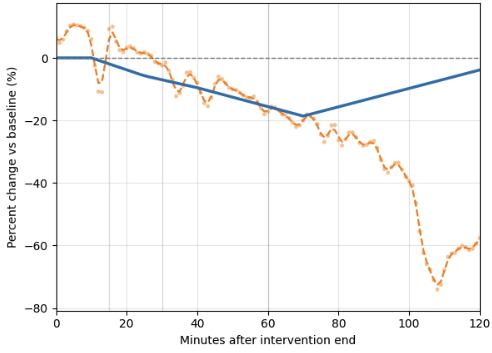}
\caption{\textbf{Example BBI post-intervention trajectories.}
\textbf{Left}: User E; \textbf{Right}: User F. Curves show the predicted median (solid blue) and the actual percent change vs.\ baseline (orange; dashed with points). The horizontal dashed line marks zero (no change). 
Across both cases the model \emph{gets the sign right} throughout most windows. Amplitudes match most closely in the first $\sim$60 minutes; beyond that, the prediction tends to under-shoot the magnitude while preserving the correct positive/negative direction (e.g., brief dip for User E around 60–80\,min; steady downward trend for User F).}
\label{fig:bbi_examples}
\end{figure}

\textbf{\emph{Trajectory-level predictions capture short-horizon dynamics before magnitude uncertainty grows.}} Our framework produces full post-intervention trajectories, not just windowed sign calls or direction predictions. For example, Fig.~\ref{fig:bbi_examples} shows this behavior for BBI trajectory prediction. The predicted median closely follows the observed post-intervention trend for approximately the first 30–60 minutes, during which autonomic adjustments are most rapid and BBI is highly informative~\citep{shaferHRVBioReview2017}. Beyond this early horizon, prediction amplitudes begin to under- or over-shoot the true curves, reflecting growing uncertainty, the natural flattening or variability of physiological recovery, and the influence of subsequent activities or events that may interact with the original intervention. Crucially, even as magnitudes drift at longer horizons, the model preserves the correct positive/negative \emph{direction} of change across windows. Together with the windowed accuracy results, these trajectory trends suggest that the framework learns physiologically plausible short-horizon dynamics while appropriately expressing uncertainty in longer-term magnitude.

\section*{Discussion and Conclusion}

\textbf{\emph{Feasibility of personalized predictions.}}
In this work, we ask whether, given the context and events of a specific day, we can predict how a particular intervention will change a person’s physiology over the next 15--120 minutes, and in what direction. The empirical results from this proof-of-concept cohort show that the model can predict short-horizon direction of change at the level of individual users and interventions, with behavior that is consistent with the established autonomic physiology literature. BBI emerges as the most informative signal across windows, RMSSD provides sharp but brief early-window information, and HR supports general trends rather than fine-grained guidance.

\textbf{\emph{From signals to personalized, intervention-level guidance and insights.}}
The combination of windowed accuracy, confusion matrices, and trajectory visualizations shows that the model makes confident directional predictions when the physiological signal is strong and abstains when it is weak or noisy and unlikely to provide actionable insight. Per-intervention heatmaps reveal stable directional signatures and heterogeneous cases where signs flip over time or differ from a user’s typical pattern. These are the settings where personalization matters most. In practice, the framework could help users and coaches identify “high-confidence” interventions that consistently move physiology in a desired direction for a given person and context, and flag interventions whose effects appear too noisy or not meaningful.

\textbf{\emph{Limitations and future directions.}}
This work serves as a proof of concept with several important limitations. The cohort is small, drawn from university settings, and monitored for only a few weeks, limiting generalizability. Intervention labels are user-entered and may be imprecise in timing and semantics. Models are trained on limited data; although we take care to avoid temporal leakage, overfitting remains a risk. Future studies should recruit larger and more diverse cohorts, extend monitoring periods, and incorporate richer contextual variables such as menstrual cycle, environmental conditions, and concurrent medications. Physiological responses to stress are further influenced by the law of initial values, baseline fitness or training status, and individual responder/non-responder phenotypes, which we do not model explicitly and which remain important directions for future work.

\FloatBarrier
\begingroup
\setlength{\textfloatsep}{6pt}
\setlength{\intextsep}{6pt}
\setlength{\abovecaptionskip}{3pt}
\setlength{\belowcaptionskip}{0pt}

\begin{figure}[!htbp]
\centering
\subfloat[\textbf{User G} — per-intervention predicted signs across windows (RMSSD, BBI, HR).]{
  \includegraphics[width=.85\linewidth,height=0.24\textheight,keepaspectratio]{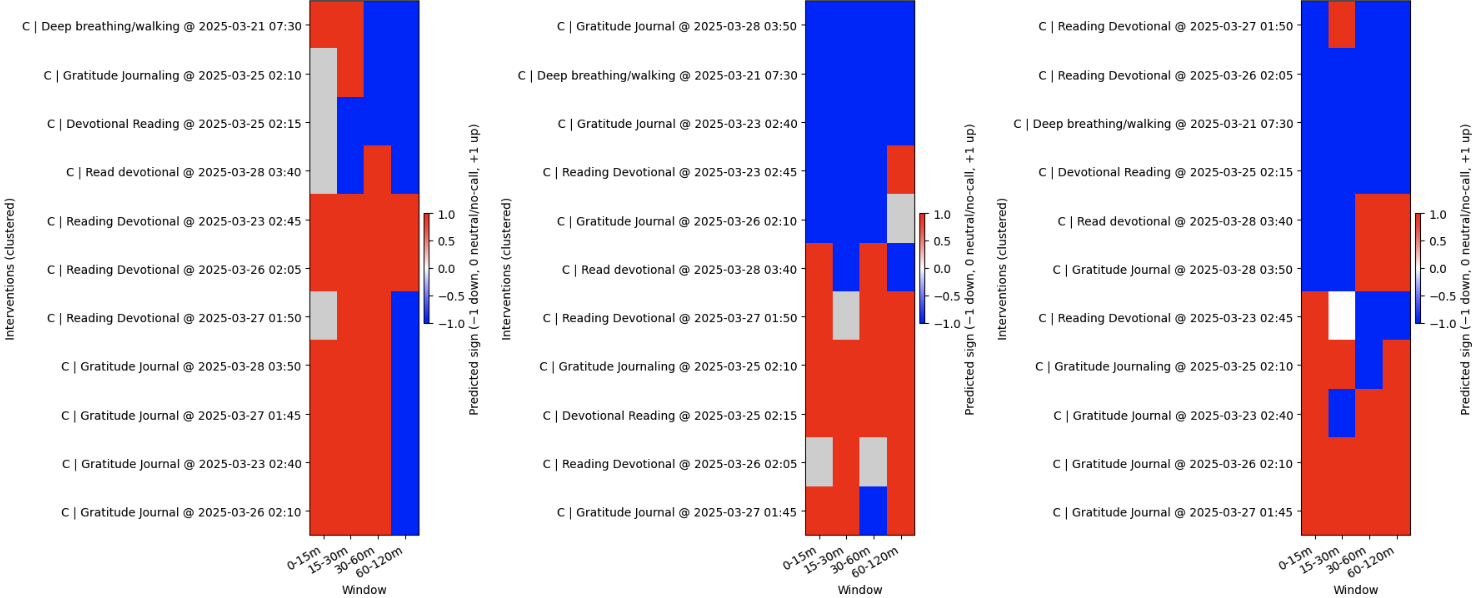}
  \label{fig:userG_heat}
}
\captionsetup{font=small}
\caption[Per-intervention sign heatmap for one user]{\textbf{Per-intervention sign heatmap for one user.}
Rows are interventions; columns are post-end windows (0–15, 15–30, 30–60, 60–120\,min).
Color key: \textcolor{red}{red} = positive sign, \textcolor{blue}{blue} = negative sign, \textcolor{gray}{gray} = neutral; lighter gray = missing.
\textbf{Takeaway:} For the participant, many interventions form contiguous vertical bands of consistent positive or negative signs across adjacent windows, indicating stable temporal response patterns, while a smaller subset show heterogeneous responses across windows.}
\label{fig:user_heatmaps}
\end{figure}
\endgroup

\clearpage
\makeatletter
\renewcommand{\@biblabel}[1]{\hfill #1.}
\bibliographystyle{unsrtnat}
\bibliography{amia}

\begin{thebibliography}{43}
\providecommand{\natexlab}[1]{#1}
\providecommand{\url}[1]{\texttt{#1}}
\expandafter\ifx\csname urlstyle\endcsname\relax
  \providecommand{\doi}[1]{doi: #1}\else
  \providecommand{\doi}{doi: \begingroup \urlstyle{rm}\Url}\fi

\bibitem[Chalmers et~al.(2021)Chalmers, Hickey, Newton, Lin, Sibbritt,
  McLachlan, Clifton-Bligh, Morley, and Lal]{chalmers2021stress}
Taryn Chalmers, Blake~Anthony Hickey, Phillip Newton, Chin-Teng Lin, David
  Sibbritt, Craig~S McLachlan, Roderick Clifton-Bligh, John Morley, and Sara
  Lal.
\newblock Stress watch: The use of heart rate and heart rate variability to
  detect stress: A pilot study using smart watch wearables.
\newblock \emph{Sensors}, 22\penalty0 (1):\penalty0 151, 2021.

\bibitem[Dunn et~al.(2021)Dunn, Kidzinski, Runge, Witt, Hicks,
  Sch{\"u}ssler-Fiorenza~Rose, Li, Bahmani, Delp, Hastie,
  et~al.]{dunn2021wearable}
Jessilyn Dunn, Lukasz Kidzinski, Ryan Runge, Daniel Witt, Jennifer~L Hicks,
  Sophia~Miryam Sch{\"u}ssler-Fiorenza~Rose, Xiao Li, Amir Bahmani, Scott~L
  Delp, Trevor Hastie, et~al.
\newblock Wearable sensors enable personalized predictions of clinical
  laboratory measurements.
\newblock \emph{Nature Medicine}, 27\penalty0 (6):\penalty0 1105--1112, 2021.

\bibitem[Hickey et~al.(2021)Hickey, Chalmers, Newton, Lin, Sibbritt, McLachlan,
  Clifton-Bligh, Morley, and Lal]{hickey2021smart}
Blake~Anthony Hickey, Taryn Chalmers, Phillip Newton, Chin-Teng Lin, David
  Sibbritt, Craig~S McLachlan, Roderick Clifton-Bligh, John Morley, and Sara
  Lal.
\newblock Smart devices and wearable technologies to detect and monitor mental
  health conditions and stress: A systematic review.
\newblock \emph{Sensors}, 21\penalty0 (10):\penalty0 3461, 2021.

\bibitem[Pakhomov et~al.(2020)Pakhomov, Thuras, Finzel, Eppel, and
  Kotlyar]{pakhomov2020using}
Serguei~VS Pakhomov, Paul~D Thuras, Raymond Finzel, Jerika Eppel, and Michael
  Kotlyar.
\newblock Using consumer-wearable technology for remote assessment of
  physiological response to stress in the naturalistic environment.
\newblock \emph{PLOS ONE}, 15\penalty0 (3):\penalty0 e0229942, 2020.

\bibitem[Smith et~al.(2020)Smith, Santoro, Moraveji, Susi, and
  Crum]{smith2020integrating}
Eric~N Smith, Erik Santoro, Neema Moraveji, Michael Susi, and Alia~J Crum.
\newblock Integrating wearables in stress management interventions: Promising
  evidence from a randomized trial.
\newblock \emph{International Journal of Stress Management}, 27\penalty0
  (2):\penalty0 172--185, 2020.

\bibitem[Pinge et~al.(2024)Pinge, Gad, Jaisighani, Ghosh, and Sen]{Pinge2024}
Ashutosh Pinge, Vivek Gad, Dhanashri Jaisighani, Soumyadeep Ghosh, and Sanchita
  Sen.
\newblock Detection and monitoring of stress using wearables: a systematic
  review.
\newblock \emph{Frontiers in Computer Science}, 6:\penalty0 1478851, 2024.
\newblock \doi{10.3389/fcomp.2024.1478851}.

\bibitem[Neupane et~al.(2024)Neupane, Saha, Ali, Hnat, Samiei, Nandugudi,
  Almeida, and Kumar]{neupane2024momentary}
Sameer Neupane, Mithun Saha, Nasir Ali, Timothy Hnat, Shahin~Alan Samiei,
  Anandatirtha Nandugudi, David~M Almeida, and Santosh Kumar.
\newblock Momentary stressor logging and reflective visualizations:
  Implications for stress management with wearables.
\newblock In \emph{Proceedings of the 2024 CHI Conference on Human Factors in
  Computing Systems}, pages 1--19, 2024.

\bibitem[Novak et~al.(2023)Novak, Robinson, Kandu{\v{c}}, Sarigiannis,
  D{\v{z}}eroski, and Kocman]{novak2023empowering}
Rok Novak, Johanna~Amalia Robinson, Tja{\v{s}}a Kandu{\v{c}}, Dimosthenis
  Sarigiannis, Sa{\v{s}}o D{\v{z}}eroski, and David Kocman.
\newblock Empowering participatory research in urban health: Wearable biometric
  and environmental sensors for activity recognition.
\newblock \emph{Sensors}, 23\penalty0 (24):\penalty0 9890, 2023.

\bibitem[Suh et~al.(2024)Suh, Howe, Lewis, Hernandez, Saha, Althoff, and
  Czerwinski]{suh2024toward}
Jina Suh, Esther Howe, Robert Lewis, Javier Hernandez, Koustuv Saha, Tim
  Althoff, and Mary Czerwinski.
\newblock Toward tailoring just-in-time adaptive intervention systems for
  workplace stress reduction: Exploratory analysis of intervention
  implementation.
\newblock \emph{JMIR Mental Health}, 11:\penalty0 e48974, 2024.

\bibitem[Comito et~al.(2021)Comito, Falcone, and
  Forestiero]{comito2021diagnosis}
Carmela Comito, Deborah Falcone, and Agostino Forestiero.
\newblock Diagnosis prediction based on similarity of patients physiological
  parameters.
\newblock In \emph{Proceedings of the 2021 IEEE/ACM International Conference on
  Advances in Social Networks Analysis and Mining}, pages 487--494, 2021.

\bibitem[Evin et~al.(2022)Evin, Hidalgo-Munoz, B{\'e}quet, Moreau, Tattegrain,
  Berthelon, Fort, and Jallais]{evin2022personality}
Morgane Evin, Antonio Hidalgo-Munoz, Adolphe~James B{\'e}quet, Fabien Moreau,
  Hel{\`e}ne Tattegrain, Catherine Berthelon, Alexandra Fort, and Christophe
  Jallais.
\newblock Personality trait prediction by machine learning using physiological
  data and driving behavior.
\newblock \emph{Machine Learning with Applications}, 9:\penalty0 100353, 2022.

\bibitem[Olyanasab and Annabestani(2024)]{olyanasab2024leveraging}
Ali Olyanasab and Mohsen Annabestani.
\newblock Leveraging machine learning for personalized wearable biomedical
  devices: A review.
\newblock \emph{Journal of Personalized Medicine}, 14\penalty0 (2):\penalty0
  203, 2024.

\bibitem[Brons et~al.(2024)]{Brons2024}
Judith Brons et~al.
\newblock Personalizing persuasive strategies in mhealth: A review.
\newblock 2024.

\bibitem[Fang et~al.(2024)]{Fang2024}
X.~Fang et~al.
\newblock Deep reinforcement learning for dynamic personalization of exercise
  goals.
\newblock 2024.

\bibitem[Sadhu et~al.(2023)]{Sadhu2023}
Arnab Sadhu et~al.
\newblock Careportal: Interactive visualization of patient-generated health
  data.
\newblock 2023.

\bibitem[Arun et~al.(2024)]{Arun2024}
S.~Arun et~al.
\newblock Remotehealthconnect: Visual analytics for remote wearable monitoring.
\newblock 2024.

\bibitem[Abdelaal et~al.(2024)]{Abdelaal2024}
Tarek Abdelaal et~al.
\newblock Explainable ai for wearable data analytics: A survey.
\newblock 2024.

\bibitem[Holzinger(2016)]{Holzinger2016}
Andreas Holzinger.
\newblock Interactive machine learning for health informatics: When do we need
  the human-in-the-loop?
\newblock In \emph{Brain Informatics and Health}, pages 1--12. Springer, 2016.

\bibitem[Esna~Ashari and Ghasemzadeh(2019)]{Ashari2019}
A.~Esna~Ashari and H.~Ghasemzadeh.
\newblock Active learning framework for wearable sensor data.
\newblock 2019.

\bibitem[Wu et~al.(2023)]{Wu2023}
X.~Wu et~al.
\newblock Integrating clinical expertise into reinforcement learning for
  treatment planning.
\newblock 2023.

\bibitem[Vaswani et~al.(2017)Vaswani, Shazeer, Parmar, Uszkoreit, Jones, Gomez,
  Kaiser, and Polosukhin]{vaswani2017attention}
Ashish Vaswani, Noam Shazeer, Niki Parmar, Jakob Uszkoreit, Llion Jones,
  Aidan~N Gomez, {\L}ukasz Kaiser, and Illia Polosukhin.
\newblock Attention is all you need.
\newblock In \emph{Advances in Neural Information Processing Systems
  (NeurIPS)}, pages 5998--6008, 2017.

\bibitem[Lim et~al.(2021)]{lim2021temporal}
Bryan Lim et~al.
\newblock Temporal fusion transformers for interpretable multi-horizon time
  series forecasting.
\newblock \emph{International Journal of Forecasting}, 37\penalty0
  (4):\penalty0 1748--1764, 2021.

\bibitem[Koenker and Bassett(1978)]{koenker1978regression}
Roger Koenker and Gilbert Bassett.
\newblock Regression quantiles.
\newblock \emph{Econometrica}, 46\penalty0 (1):\penalty0 33--50, 1978.

\bibitem[Geifman and El-Yaniv(2017)]{geifman2017selective}
Yonatan Geifman and Ran El-Yaniv.
\newblock Selective classification for deep neural networks.
\newblock In \emph{Advances in Neural Information Processing Systems
  (NeurIPS)}, 2017.

\bibitem[Shaffer and Ginsberg(2017{\natexlab{a}})]{Shaffer2017Overview}
Fred Shaffer and Jay~P Ginsberg.
\newblock An overview of heart rate variability metrics and norms.
\newblock \emph{Frontiers in Public Health}, 5:\penalty0 258,
  2017{\natexlab{a}}.
\newblock \doi{10.3389/fpubh.2017.00258}.

\bibitem[Laborde et~al.(2017{\natexlab{a}})Laborde, Mosley, and
  Thayer]{Laborde2017HRVGuidelines}
Sylvain Laborde, Emma Mosley, and Julian~F. Thayer.
\newblock Heart rate variability and cardiac vagal tone in psychophysiological
  research: Recommendations for experiment planning, data analysis, and data
  reporting.
\newblock \emph{Frontiers in Psychology}, 8:\penalty0 213, 2017{\natexlab{a}}.

\bibitem[Kim et~al.(2018{\natexlab{a}})Kim, Cheon, Bai, Lee, and
  Koo]{Kim2018HRVReview}
H.-G. Kim, E.-J. Cheon, D.-S. Bai, Y.~H. Lee, and B.-H. Koo.
\newblock Heart rate variability and its application in clinical and exercise
  physiology.
\newblock \emph{Frontiers in Physiology}, 9:\penalty0 188, 2018{\natexlab{a}}.

\bibitem[Finseth et~al.(2023)Finseth, Dorneich, Vardeman, Keren, and
  Franke]{finseth2023real}
Tor~T Finseth, Michael~C Dorneich, Stephen Vardeman, Nir Keren, and Warren~D
  Franke.
\newblock Real-time personalized physiologically based stress detection for
  hazardous operations.
\newblock \emph{IEEE Access}, 11:\penalty0 25431--25454, 2023.

\bibitem[Jaafar and Chung~Xian(2021)]{jaafar2021analysis}
Rosmina Jaafar and Onn Chung~Xian.
\newblock Analysis of heart rate variability using wearable device.
\newblock In \emph{Computational Science and Technology: 7th ICCST 2020,
  Pattaya, Thailand, 29--30 August, 2020}, pages 453--461. Springer, 2021.

\bibitem[Sheridan et~al.(2020)Sheridan, Dehart, Lin, Sabbaj, and
  Baker]{sheridan2020heart}
David~C Sheridan, Ryan Dehart, Amber Lin, Michael Sabbaj, and Steven~D Baker.
\newblock Heart rate variability analysis: How much artifact can we remove?
\newblock \emph{Psychiatry Investigation}, 17\penalty0 (9):\penalty0 960, 2020.

\bibitem[Dalmeida and Masala(2021)]{Dalmeida2021HRVStress}
Kathryn~M. Dalmeida and Giovanni~Luca Masala.
\newblock Hrv features as viable physiological markers for stress detection
  using wearable devices.
\newblock \emph{Sensors}, 21\penalty0 (8):\penalty0 2873, 2021.
\newblock \doi{10.3390/s21082873}.

\bibitem[Tervonen et~al.(2021)Tervonen, Pettersson, and
  M{\"a}ntyj{\"a}rvi]{Tervonen2021UltraShort}
Jaakko Tervonen, Kati Pettersson, and Jani M{\"a}ntyj{\"a}rvi.
\newblock Ultra-short window length and feature importance analysis for
  cognitive load detection from wearable sensors.
\newblock \emph{Electronics}, 10\penalty0 (5):\penalty0 613, 2021.
\newblock \doi{10.3390/electronics10050613}.

\bibitem[Velmovitsky et~al.(2023)Velmovitsky, Lotto, Alencar, Leatherdale,
  Cowan, and Morita]{Velmovitsky2023AppleWatch}
Pedro~Elkind Velmovitsky, Matheus Lotto, Paulo Alencar, Scott~T. Leatherdale,
  Donald Cowan, and Plinio~Pelegrini Morita.
\newblock Can heart rate variability data from the apple watch
  electrocardiogram quantify stress?
\newblock \emph{Frontiers in Public Health}, 11:\penalty0 1178491, 2023.
\newblock \doi{10.3389/fpubh.2023.1178491}.

\bibitem[Liu et~al.(2023)Liu, Jiao, Du, Zhang, Chen, Xu, and
  Jiang]{Liu2023DriverStress}
Kun Liu, Yubo Jiao, Congcong Du, Xiaoming Zhang, Xiaoyu Chen, Fang Xu, and
  Chaozhe Jiang.
\newblock Driver stress detection using ultra-short-term hrv analysis under
  real world driving conditions.
\newblock \emph{Entropy}, 25\penalty0 (2):\penalty0 194, 2023.
\newblock \doi{10.3390/e25020194}.

\bibitem[Darwish et~al.(2025)Darwish, Rehman, Sadek, Salem, Kareem, and
  Mahmoud]{Darwish2025ThreeStage}
Basil~A. Darwish, Shafiq~Ul Rehman, Ibrahim Sadek, Nancy~M. Salem, Ghada
  Kareem, and Lamees~N. Mahmoud.
\newblock From lab to real-life: A three-stage validation of wearable
  technology for stress monitoring.
\newblock \emph{MethodsX}, 14:\penalty0 103205, 2025.
\newblock \doi{10.1016/j.mex.2025.103205}.

\bibitem[Shaffer and Ginsberg(2017{\natexlab{b}})]{shaferHRVBioReview2017}
Fred Shaffer and JP~Ginsberg.
\newblock An overview of heart rate variability metrics and norms.
\newblock \emph{Frontiers in Public Health}, 5:\penalty0 258,
  2017{\natexlab{b}}.

\bibitem[Kim et~al.(2018{\natexlab{b}})Kim, Cheon, and Bai]{kimHRVReview2018}
Hyun-Joong Kim, Eunkyoung Cheon, and Dae-Woong Bai.
\newblock The relationship between heart rate variability and stress.
\newblock \emph{Psychiatry Investigation}, 15\penalty0 (3):\penalty0 235--245,
  2018{\natexlab{b}}.

\bibitem[Laborde et~al.(2017{\natexlab{b}})Laborde, Mosley, and
  Thayer]{laborde2017hrv}
Sylvain Laborde, Emma Mosley, and Julian~F Thayer.
\newblock Heart rate variability and cardiac vagal tone in psychophysiological
  research--recommendations for experiment planning, data analysis, and data
  reporting.
\newblock \emph{Frontiers in Psychology}, 8:\penalty0 213, 2017{\natexlab{b}}.
\newblock \doi{10.3389/fpsyg.2017.00213}.

\bibitem[Stanley et~al.(2013)Stanley, Peake, and Buchheit]{stanleyHRvsHRV2013}
Jamie Stanley, Jonathan~M Peake, and Martin Buchheit.
\newblock Recovery from exercise: a brief review focusing on heart rate and
  heart rate variability.
\newblock \emph{Journal of Strength and Conditioning Research}, 27\penalty0
  (11):\penalty0 3174--3182, 2013.

\bibitem[Buchheit(2014)]{buchheit2014sensitivity}
Martin Buchheit.
\newblock Sensitivity of postexercise heart rate variability to training.
\newblock \emph{Sports Medicine}, 44\penalty0 (5):\penalty0 569--581, 2014.
\newblock \doi{10.1007/s40279-013-0130-8}.

\bibitem[Laborde et~al.(2017{\natexlab{c}})Laborde, Mosley, and
  Thayer]{labordeHRVGuidelines2017}
Sylvain Laborde, Emma Mosley, and Julian Thayer.
\newblock Heart rate variability and cardiac vagal tone in psychophysiological
  research—recommendations for experiment planning, data analysis, and
  reporting.
\newblock \emph{Frontiers in Psychology}, 8:\penalty0 213, 2017{\natexlab{c}}.

\bibitem[Järvelä(2021)]{jarvela_earlywindow2021}
Matti et~al. Järvelä.
\newblock Short-term autonomic responses following physical activity:
  Characterizing early recovery windows using hrv metrics.
\newblock \emph{European Journal of Applied Physiology}, pages 1--12, 2021.

\bibitem[Cole et~al.(1996)Cole, Blackstone, Pashkow, Snader, and
  Lauer]{bruceHRRecovery1996}
Charles Cole, Eugene Blackstone, Fred Pashkow, Charles Snader, and Michael
  Lauer.
\newblock Heart rate recovery: Clinical implications and physiology.
\newblock \emph{Journal of the American College of Cardiology}, 28:\penalty0
  1527--1533, 1996.

\end{thebibliography}

\end{document}